# Evolving TSP heuristics using Multi Expression Programming

**Mihai Oltean, D. Dumitrescu**

Department of Computer Science,
Faculty of Mathematics and Computer Science,
Babeş-Bolyai University, Kogălniceanu 1,
Cluj-Napoca, 3400, Romania.
email: {moltean, ddumitr}@nessie.cs.ubbcluj.ro

**Abstract.** Multi Expression Programming (MEP) is an evolutionary technique that may be used for solving computationally difficult problems. MEP uses a linear solution representation. Each MEP individual is a string encoding complex expressions (computer programs). A MEP individual may encode multiple solutions of the current problem. In this paper MEP is used for evolving a Traveling Salesman Problem (TSP) heuristic for graphs satisfying triangle inequality. Evolved MEP heuristic is compared with Nearest Neighbor Heuristic (NN) and Minimum Spanning Tree Heuristic (MST) on some difficult problems in TSPLIB. For most of the considered problems the evolved MEP heuristic outperforms NN and MST. The obtained algorithm was tested against some problems in TSPLIB. The results emphasizes that evolved MEP heuristic is a powerful tool for solving difficult TSP instances.

## 1. Introduction

In [12, 13, 14] a new evolutionary paradigm called *Multi Expression Programming* (MEP)[1] has been proposed. MEP may be considered as an alternative to standard Genetic Programming technique [8]. MEP uses a linear solution representation. Each MEP individual is a string encoding complex expressions (computer programs). A MEP individual may encode multiple solutions of the current problem. Usually the best solution is chosen for fitness assignment purposes.

One of the most important applications of MEP is discovering heuristics for solving computationally difficult (mainly NP–Complete) problems. Instead of searching the solution of a particular problem the MEP aim is to discover a heuristic that solves the entire class of instances for a given problem.

In this paper MEP technique is used for discovering TSP heuristics for graphs satisfying triangle inequality (TI graphs). This option was chosen due to the existence of a big number of real-world applications implying TI graphs (e.g. plains, trains and vehicles routes). MEP technique is used to learn a path function $f$ that is used for evaluating the reachable nodes. This function serves as a heuristic for detecting the optimum path.

In the proposed approach the TSP path starts with a randomly selected node of the graph. Each node reachable from the current node in one step is evaluated using the function (computer program) $f$ evolved by MEP algorithm. The best node is added to the already detected path. The algorithm stops when the path contains all graph nodes.

---

[1] MEP source code is available at the address www.mep.cs.ubbcluj.ro.

MEP learning process for TSP has a remarkable quality: the evolved (learned) heuristic works very well for data sets much larger than the training set. For MEP training stage graphs having 3 to 50 nodes are considered. Evolved MEP function was tested and performs well for graphs having maximum 1000 nodes.

Evolved function *f* is compared with some well known heuristics. Numerical experiments emphasize that (for considered examples) MEP function outperforms dedicated heuristics.

## 2. MEP Technique

MEP uses a linear solution representation and a special phenotypic transcription model. A MEP chromosome usually encodes several expressions (computer programs). The ability of MEP chromosome to encode several syntactically correct expressions is called *strong implicit parallelism*.

### 2.1. MEP Algorithm

Standard MEP algorithm starts with a randomly generated population of individuals.

A fixed number of the high fit individuals enter in the next generation (elitism). The mating pool is filled using binary tournament selection. Individuals from mating pool are randomly paired and recombined. By recombination of two parents two offspring are obtained. The offspring are mutated and enter the next generation.

### 2.2. MEP Representation

MEP genes are substrings of variable length. Number of genes in a chromosome is constant and it represents *chromosome length*. Each gene encodes a terminal or a function symbol. A gene encoding a function includes pointers towards genes containing the function arguments. Function parameters always have indices of lower values than the position of that function symbol itself in chromosome.

Proposed representation ensures no cycle arises when the chromosome is decoded (phenotypically transcripted). According to the proposed representation scheme the first symbol of the chromosome must be a terminal symbol. In this way only syntactically correct programs are generated by MEP technique.

Let $T = \{a, b, c, d\}$ be the set of terminal symbols and $F = \{+, *\}$ be the set of function symbols. Consider as an example the MEP chromosome *C* given below:

1: *a*
2: *b*
3: + 1, 2
4: *c*
5: *d*
6: * 4, 5

***Remark***: Numbers on the left positions stand for gene labels or addresses. Actually labels do not belong to the chromosome, but they are provided for explanation purposes only.

**2.3. MEP phenotypic transcription**

MEP chromosomes are read downstream starting with the first position. A terminal symbol specifies a simple expression. A function symbol specifies a complex expression (formed by connecting the operands specified by the argument positions with the current function symbol).

Consider the chromosome *C* specified above (section 2.2). Chromosome *C* is not able to encode a unique expression that involves all of the genes. But *C* encodes the expressions:

$E_1 = a$,
$E_2 = b$,
$E_3 = a + b$,
$E_4 = c$,
$E_5 = d$,
$E_6 = c + d$.

Each MEP chromosome is allowed to encode a number of expressions equal to the chromosome length (number of genes). Expression associated to each chromosome position is obtained by interpreting the respective gene.

**2.4. Selection and search operators**

Within MEP technique binary tournament [Goldberg] selection is used. Search operators are recombination and mutation. These possible operators are defined to preserve the chromosome structure. All offspring describe syntactically correct expressions.

**2.4.1. Recombination**

Three variants of recombination operator have been considered and tested within our MEP implementation: one–point crossover, two–point crossover and uniform crossover. These operators are simple versions of standard binary crossover operators (see [4], [6]). Two–point crossover seems to work best with MEP ([12]) and it will be used in all experiments considered in this paper.

**2.4.2. Mutation**

Mutation operator may be applied to each chromosome gene. A mutation probability ($p_m$) is considered when applying mutation operator.

By mutation some symbols in chromosome are changed. To preserve the chromosome structure its first gene must encode, also after mutation, a terminal symbol. For other genes there is no restriction in symbols changing.

If the gene selected for mutation encodes a terminal symbol, this symbol may be changed into another terminal symbol or into a function symbol. In the last case the positions (addresses) indicating the function arguments are randomly generated.

If the mutating gene encodes a function, then the gene may be mutated into a terminal symbol or into another function (i.e. function symbol and pointers towards arguments).

## 3. TSP problem with triangle inequality

TSP problem for TI graphs (i.e. satisfying triangle inequality) is stated as follows. Consider a set $C = \{c_0, c_1, \ldots, c_{N-1}\}$ of cities, and a distance $d(c_i, c_j) \in Z^+$ for each pair $c_i, c_j \in C$, $d(c_i, c_j) = d(c_j, c_i)$, and for each three cities $c_i, c_j, c_k \in C$, $d(c_i, c_j) \leq d(c_i, c_k) + d(c_k, c_j)$. The tour $<c_{\pi(0)}, c_{\pi(1)}, \ldots, c_{\pi(N-1)}>$ of all cities in $C$ having minimum length is needed ([1], [3])

TSP problem with triangle inequality is an NP–complete problem [7]. No polynomial time algorithm for solving TSP problem is known.
Several heuristics for solving TSP problem have been proposed. The most important are Nearest Neighbor ([3], [7]) and the Minimum Spanning Tree ([3]).

## 4. Evolving Heuristics for TSP

In this section we address the problem of discovering heuristics that can solve TSP rather than solving a particular instance of the problem.

MEP technique is used for evolving a path function $f$ that gives a way to choose graph vertices in order to obtain a Hamiltonian cycle. The fitness is assigned to a function $f$ in the current population by applying $f$ on several randomly chosen graphs (training set) and evaluating the results.

Evolved path function may be used for solving particular instances of TSP. For each problem the graph nodes are evaluated using the path function $f$ and are added one by one to the already build path.

The algorithm for TSP using evolved path function $f$ may be described as follows:

$S_1$. Let $c_{\pi(0)} = c_0$ {the path starts with the node $c_0$}
$S_2$. $k = 1$;
$S_3$. **while** $k < N - 1$ **do**
$S_4$.   Using function $f$ select $c_{\pi(k+1)}$ – the next node of the path
$S_5$.   Add $c_{\pi(k+1)}$ to the already built path.
$S_6$.   $k = k + 1$;
$S_7$. **endwhile**

$S_4$ is the key step of this algorithm. The procedure that selects the next node of the path in an optimal way uses the function $f$ evolved by the MEP technique as described in sections 4.1 and 4.2.

### 4.1. Terminal and Function Symbols for Evolving Heuristic Function $f$

Path function $f$ has to use (as input) some information about already build path and some information about unvisited nodes.

A natural way for defining the set of terminals is to consider the terminals as representing the distances between nodes. Therefore we have:
$T = \{d_{i,j}, | 0 \leq i \leq N - 1, 0 \leq j \leq N - 1\}$.

But this approach leads to some difficulties when applied for graphs having different number of nodes. To avoid this difficulty, we consider a special terminal set which is independent with respect to the number of graph nodes.

Let us denote by $y_1$ the last visited node (current node). We have to select the next node to be added to the path. In this respect all unvisited nodes are considered. Let us denote by $y_2$ the next node to be visited.

For evolving path function $f$ we consider a set $T$ of terminals involving the following elements:

(i) $d\_y_1\_y_2$ – distance between the graph nodes $y_1$ and $y_2$,
(ii) $min\_g\_y_1$ ($min\_g\_y_2$) – the minimum distance from the nodes $y_1$ ($y_2$) to unvisited nodes,
(iii) $sum\_g\_y_1$ ($sum\_g\_y_2$) – the sum of all distances between nodes $y_1$ ($y_2$) and unvisited nodes,
(iv) $prod\_g\_y_1$ ($prod\_g\_y_2$) – the product of all distances between nodes $y_1$ ($y_2$) and unvisited nodes,
(v) $max\_g\_y_1$ ($max\_g\_y_2$) – the maximum distance from the nodes $y_1$ ($y_2$) to unvisited nodes,
(vi) *length* – the length of the already built path.

The set $T$ of terminals (function variables) is thus:

$T = \{d\_y_1\_y_2, min\_g\_y_1, min\_g\_y_2, max\_g\_y_1, max\_g\_y_2, sum\_g\_y_1, sum\_g\_y_2, prod\_g\_y_1, prod\_g\_y_2, length\}$.

Let us remark that members of $T$ are not actual terminals (in the standard acceptation). For this reason we may call members of $T$ as *instantiated* (or *intermediate*) *nonterminals*.

Set $T$ of terminals is chosen in such way to be independent of the number of graph nodes. This choice confers flexibility and robustness to the evolved heuristic.

For evolving a MEP function for TSP problem we may consider the following set of function symbols: $F = \{+, -, /, *, cos, sin, min, max\}$.

The node $y_2$ that generates the lowest output of evolved function $f$ is chosen to be the next node of the path. Ties are solved arbitrarily. For instance we may consider the node with the lowest index is selected.

**Example**

Consider the MEP linear structure:
1: $d\_y_1\_y_2$
2: $min\_g\_y_1$
3: + 1, 2
4: $sum\_g\_y_2$
5: * 2, 4

This MEP individual encodes the path functions $f_1, f_2, f_3, f_4, f_5$ given by:
$f_1 = d\_y_1\_y_2$,
$f_2 = min\_g\_y_1$,
$f_3 = d\_y_1\_y_2 + min\_g\_y_1$,
$f_4 = sum\_g\_y_2$,
$f_5 = min\_g\_y_1 * sum\_g\_y_2$.

### 4.2. Fitness assignment

In order to obtain a good heuristic we have to train the path function $f$ using several graphs. The training graphs are randomly generated at the beginning of the search process and remain unchanged during the search process. To avoid overfitting (see [15]), another set of randomly generated graphs (validation set) is considered. After each generation the quality of the best-so-far individual is calculated using the validation set in order to check its generalization ability during training. At the end of the search process, the function with the highest quality is supplied as the program output.

The fitness (quality) of a detected path function $f$ is defined as the sum of the TSP path length of graphs in the training set. Thus the fitness is to be minimized.

### 4.3. A Numerical Experiment

In this experiment we evolve a heuristic for solving TSP problem.

Let us denote by $G_k$ the set of class of TI graphs having maximum $k$ nodes. MEP algorithm considers the class $G_{50}$ (i.e. graphs having 3 to 50 nodes) for training and the class $G_{100}$ for validation. Evolved path function was tested for graphs in the class $G_{1000}$ (i.e. graphs having maxim 1000 nodes). MEP algorithm parameters are given in Table 1.

| Population size | 300 |
|---|---|
| Number of generations | 100 |
| Chromosome length | 40 genes |
| Mutation probability | 0.1 |
| Crossover type | One-Crossover-Point |
| Crossover probability | 0.9 |
| Training set size | 30 |
| Maximum number of nodes in training set | 50 |
| Validation set size | 20 |
| Maximum number of nodes in validation set | 100 |

**Table 1**. MEP algorithm parameters for evolving a heuristic for TSP with triangle inequality.

The evolution of the best individual fitness and the average fitness of the best individuals over 30 runs are depicted in Figure 1.

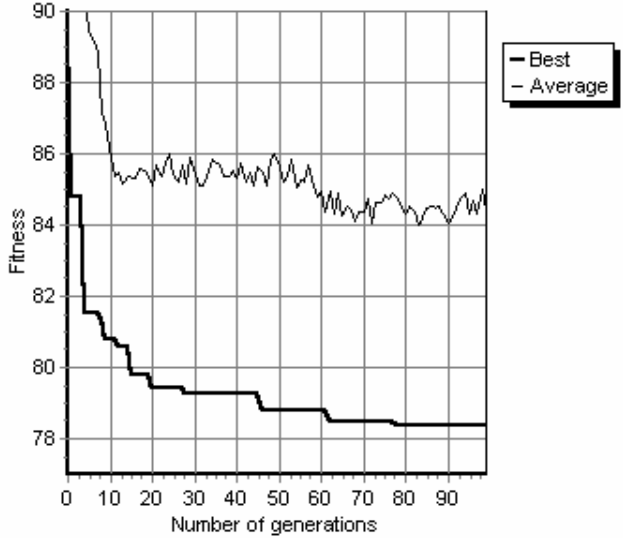

**Figure 1**. The fitness evolution of the best individual in the best run and the average fitness of the best individuals over 30 runs.

A path function evolved by the MEP algorithm is:
$f = (sum\_g(y2)) * (d\_y_1\_y_2 - (max(d\_y_1\_y_2, max\_g(y_1))) + d\_y_1\_y_2)$.

Heuristic function $f$ that is evolved by MEP technique is directly used for building the optimum path. The corresponding learning process has a remarkable quality: the evolved (learned) heuristic works very well on data sets significantly larger than the training set. In our example the training set $G_{50}$ is significantly smaller than the set $G_{1000}$ used for testing.

## 5. Assessing the Performance of the Evolved MEP Heuristic

In this section the performance of evolved MEP heuristic, NN and MST are compared. In the first experiment we compare the considered algorithms on some randomly generated graphs. In the second experiment the heuristics are compared against several difficult problems in TSPLIB [16].

### 5.1. Experiment 1

In this experiment we provide a direct comparison of the evolved MEP heuristic, NN and MST. The considered heuristics are tested for randomly generated graphs satisfying triangle inequality.

Evolved heuristic was tested for different graphs from the classes $G_{200}$, $G_{500}$ and $G_{1000}$. For each graph class 1000 graphs satisfying triangle inequality have been randomly generated. These graphs have been considered for experiments with evolved MEP heuristic, NN and MST.

Performance of evolved MEP heuristic, NN and MST are depicted in Table 2

**Table 2**. Evolved MEP heuristic vs. NN, MST. For each graph class we present the number of graphs for which evolved MEP heuristic generates a cycle shorter than the cycle obtained by the algorithm MST and NN.

| Graphs types | MST | NN |
|---|---|---|
| $G_{200}$ | 953 | 800 |
| $G_{500}$ | 974 | 906 |
| $G_{1000}$ | 990 | 948 |

Results obtained emphasizes that evolved MEP heuristic outperforms NN and MST algorithms on random graphs.

### 5.2. Experiment 2

To obtain a stronger evidence of the results above we test the performance of the considered heuristics against some difficult problems in TSPLIB. The results are presented in Table 3.

**Table 3**. The performance of evolved MEP heuristic, NN and MST on some problems in TSPLIB. *Length* is the length of the TSP path obtained with one of the considered heuristics. Error is calculated as (*Length – Shortest_Length*)/ *Shortest_Length* * 100. Each node of the graph has been considered as the first node of the path.

| **Problem** | **MEP** | | **NN** | | **MST** | |
|---|---|---|---|---|---|---|
| | Length | Error (%) | Length | Error (%) | Length | Error (%) |
| a280 | 2858.86 | 10.85 | 3084.22 | 19.58976 | 3475.23 | 34.75 |
| att48 | 37188.2 | 10.93 | 39236.9 | 17.04227 | 43955.8 | 31.11 |
| berlin52 | 7672.1 | 1.72 | 8182.19 | 8.488332 | 10403.9 | 37.94 |
| bier127 | 134945 | 14.08 | 127954 | 8.177068 | 152747 | 29.13 |
| ch130 | 6558.03 | 7.33 | 7198.74 | 17.81899 | 8276.51 | 35.45 |
| ch150 | 7104.03 | 8.82 | 7078.44 | 8.431985 | 9142.99 | 40.05 |
| d198 | 17780.7 | 12.67 | 17575.1 | 11.37579 | 17957.6 | 13.79 |
| d493 | 43071.3 | 23.05 | 41167 | 17.61328 | 41846.6 | 19.55 |
| d657 | 56965.6 | 16.46 | 60398.7 | 23.48442 | 63044.2 | 28.89 |
| eil101 | 685.013 | 8.9 | 753.044 | 19.72083 | 846.116 | 34.51 |
| eil51 | 441.969 | 3.74 | 505.298 | 18.61455 | 605.049 | 42.03 |
| eil76 | 564.179 | 4.86 | 612.656 | 13.87658 | 739.229 | 37.4 |
| fl417 | 13933.8 | 17.47 | 13828.2 | 16.58545 | 16113.2 | 35.85 |
| gil262 | 2659.17 | 11.82 | 2799.49 | 17.72456 | 3340.84 | 40.48 |
| kroA150 | 28376.3 | 6.98 | 31482 | 18.6925 | 38754.8 | 46.11 |
| kroA200 | 32040.3 | 9.09 | 34547.7 | 17.63722 | 40204.1 | 36.89 |
| kroB100 | 24801 | 12.01 | 25883 | 16.90077 | 28803.5 | 30.09 |
| kroB200 | 33267.4 | 13.01 | 35592.4 | 20.91042 | 40619.9 | 37.98 |
| lin105 | 15133.2 | 5.24 | 16939.4 | 17.80652 | 18855.6 | 31.13 |
| lin318 | 46203.4 | 9.93 | 49215.6 | 17.09915 | 60964.8 | 45.05 |
| pcb442 | 56948.3 | 12.15 | 57856.3 | 13.9397 | 73580.1 | 44.9 |
| pr226 | 84937.8 | 5.68 | 92905.1 | 15.59818 | 111998 | 39.35 |
| pr264 | 55827.1 | 13.61 | 54124.5 | 10.15468 | 65486.5 | 33.27 |
| rat195 | 2473.49 | 6.47 | 2560.62 | 10.22901 | 2979.64 | 28.26 |
| rat575 | 7573.6 | 11.82 | 7914.2 | 16.84925 | 9423.4 | 39.13 |
| rat783 | 9982.96 | 13.36 | 10836.6 | 23.05928 | 11990.5 | 36.16 |
| rd400 | 16973.3 | 11.07 | 18303.3 | 19.77816 | 20962 | 37.17 |

| | | | | | |
|---|---|---|---|---|---|
| ts225 | 136069 | 7.44 | 140485 | 10.92994 | 187246 | 47.85 |
| u574 | 43095.6 | 16.77 | 44605.1 | 20.86465 | 50066 | 35.66 |
| u724 | 46545.7 | 11.06 | 50731.4 | 21.04844 | 60098.9 | 43.39 |

From Table 3 we can see that evolved MEP heuristic performs better than NN and MST on most of the considered problems. Only for five problems (bier127, ch150, d198, d493, fl417) NN performs better than evolved MEP heuristic. MST does not perform better than evolved MEP heuristic for no problem. The highest error obtained by the evolved MEP heuristic is 23.05 (the problem d493) while the highest error obtained by NN is 23.45 (the problem rd400). The lowest error obtained with MEP is 1.72 (problem berlin52) while the lowest error obtained by NN is 8.17 (problem bier127). The mean of errors for all considered problems is 10.61 (for evolved MEP heuristic) 16.33 (for NN heuristic) and 35.77 (for MST heuristic).

## 6. Conclusions and Further Work

MEP technique is used to evolve heuristics for solving TSP problems. Experimental results emphasizes that evolved heuristic outperforms some well known dedicated heuristics.

Moreover improvement of MEP results could be realized by allowing more function symbols to appear in the MEP chromosome. Further research will focus on using MEP for discovering better heuristics for solving TSP.

Further improvement may be obtained by increasing the chromosome length. In this case the complexity of the evolved formula could increases, but the performances of the obtained heuristic could be significantly better.

Evolving functions that outperform other dedicated heuristics would be of great practical interest. In this way computer programs that are hard to implement could be simulated by simple functions.